\documentclass[runningheads]{llncs}
\usepackage{graphicx}
\usepackage{xcolor} 
\usepackage{amsmath}
\usepackage{amssymb}
\DeclareMathOperator{\E}{\mathbb{E}}

\let\conjugatet\overline
\DeclareMathAlphabet{\pazocal}{OMS}{zplm}{m}{n}
\usepackage{algorithm,algorithmic,multicol}

\usepackage{xr}
\begin{document}
\title{Conditional WGANs with Adaptive Gradient Balancing for Sparse MRI Reconstruction}
\titlerunning{cWGANs with AGB for Sparse MRI Reconstruction}
%
\author{Itzik Malkiel\inst{1,2} \and
Sangtae Ahn\inst{1} \and
Valentina Taviani\inst{3} \and
Anne Menini\inst{3} \and
Lior Wolf\inst{2} \and
Christopher J. Hardy\inst{1}}
\authorrunning{I. Malkiel et al.}
%
\institute{GE Global Research \and School of Computer Science, Tel Aviv University, Israel \and GE Healthcare} %
\maketitle              
\begin{abstract}
Recent sparse  MRI reconstruction models have used Deep Neural Networks (DNNs) to reconstruct relatively high-quality images from highly undersampled $k$-space data, enabling much faster MRI scanning. However, these techniques sometimes struggle to reconstruct sharp images that preserve fine detail while maintaining a natural appearance. 

In this work, we enhance the image quality by using a Conditional Wasserstein Generative Adversarial Network combined with a novel Adaptive Gradient Balancing technique that stabilizes the training and minimizes the degree of artifacts, while maintaining a high-quality reconstruction that produces sharper images than other techniques.

\end{abstract}

\section{Introduction}
MRI data acquisition is inherently slow, and can often exceed 30 minutes. 
One way to accelerate MR scanning is undersampling the $k$-space, i.e., reducing the number of $k$-space traversals by a factor $R$, and accelerating the scan proportionately. Reconstruction is then performed by using parallel imaging (PI) or compressed sensing (CS) techniques.

More recently, Deep Neural Networks (DNNs) have been used to push $R$ values even higher \cite{8hammernik2018learning,9schlemper2018deep,6zhu2018image}. Among the most promising Deep Learning (DL) techniques, the unrolled iterative networks (also called cascading network) have emerged as a leading powerful method~\cite{8hammernik2018learning,9schlemper2018deep}. Inspired by CS, this technique uses a DNN composed of a sequence of iterations that include  data-consistency and convolutional units. The data-consistency units utilize the acquired $k$-space lines as a prior that keeps the network from drifting away from the acquired data, and the convolutional layers are trained to regularize the reconstruction.

As with other image generation problems, using a naive pixel-wise distance for training DL-based sparse MRI reconstruction models can result in image blurring and unrealistic appearance. In a clinical setting, avoidance of blurring can be crucial for proper diagnosis.  Recently, Generative Adversarial Networks (GANs) have been used to promote the naturalness of MRI reconstructions\cite{14hammernik2018variational,12mardani2019deep,13yang2018dagan}. In our work, we harness the power of conditional Wasserstein GANs (cWGANs) to further improve image quality.

The main contributions of this paper are as follows: (1) We propose a cWGAN method for sparse MRI reconstruction, in which both the generator and discriminator are conditioned using the acquired undersampled data. (2) We introduce a novel training algorithm called Adaptive Gradient Balancing (AGB) which balances the losses in multi-term adversarial objectives. 
(3) We provide an extensive comparison between different models and training techniques. In particular, we report results of four different techniques - an unrolled iterative network, a WGAN based network, a cWGAN network and a cWGAN network trained with our AGB.
(4) We propose and evaluate a novel Densely Connected Iterative Network (DCI-Net) for sparse  MRI reconstruction, which is inspired by Dense-Nets \cite{10huang2017densely}. (5) We are the first to adopt the Fr\'echet Inception Distance as a score metric for sparse MRI reconstruction. 

\noindent{\bf Related work~~}
DL-based sparse MRI reconstruction has attracted considerable attention recently. Schlemper et al.~\cite{9schlemper2018deep} used a cascade of CNNs optimized to minimize a pixel-wise distance. Hammernik et al. proposed Variational Networks (VN) for solving MRI-sparse reconstruction: first, a VN that minimizes a pixel-wise loss~\cite{8hammernik2018learning}, then a GAN-based VN ~\cite{14hammernik2018variational} to bear on the blurring artifacts. Mardani et al.~\cite{12mardani2019deep} proposed a GAN-based model that uses a deep residual network as a generator. Yang et al.~\cite{13yang2018dagan} introduced a GAN-based model trained to optimize a mixture of a pixel-wise loss, a perceptual loss and a GAN loss which conditions only the generator input. Yang et al. reported that a GAN-based model without perceptual loss, generates unrealistic jagged artifacts.

\section{Problem Formulation}
Let $k \in \mathbb{C}^{H \times W}$ be the $k$-space signal acquired by an MRI scanner. For a single-coil receiver, an image $m \in \mathbb{C}^{H \times W}$ can be estimated by performing an inverse Fourier transform $m=\mathcal{F}^{-1}(k)$. In multi-coil MRI, an array of $N$ coils acquire $N$ different 2D $k$-space measurements of the same object $K = \left\{ k_i | k_i \in \mathbb{C}^{H \times W}, i = 1\dots N \right\}$. Each coil $C_i$, positioned at a different location, is typically highly sensitive in one region of space. This position-dependent sensitivity can be represented by a complex-valued coil sensitivity map in real space, $S = \left\{ s_i | s_i \in \mathbb{C}^{H \times W}, i = 1\dots N \right\}$.

During reconstruction, the images from each coil are combined into a fully-sampled image $m_f = \pazocal{R}\left(K,S\right)$, where $\pazocal{R}$ is a reconstruction function $\pazocal{R}\left(K,S\right) = \sum\limits_{i=1}^N \conjugatet{s_i} \odot \mathcal{F}^{-1}(k_i)$ and $\conjugatet{s_i}$ is the complex conjugate of the sensitivity map of coil $C_i$. To accelerate imaging, a binary sampling pattern $M$ is used to undersample each coil's $k$-space signal for each slice. The undersampled $k$-space signal, denoted by $K_u$, can be written as $K_u = M \odot K$. The undersampled zero-filled image  $m_z$ can be calculated by: $m_z = \pazocal{R}\left(K_u, M\right)$. The learning task is to find a reconstruction function $G^{*}$ that minimizes an expected loss function $\pazocal{L}$ (Sec.~\ref{sec:Objective}) over a population of scans: $G^{*} = \arg\min_{G}\mathbb{E}_{\left(K,M\right)}\left[\pazocal{L}\left(G \left( K_u , M\right), m_f\right) \right]$. For a given $G^*$, $K_u$ and $M$, we will denote by $m_g$ the generated image $m_g := G^{*}\left(K_u,M\right)$.

\section{Method}

\begin{figure}[t]
\includegraphics[width=0.8\linewidth]{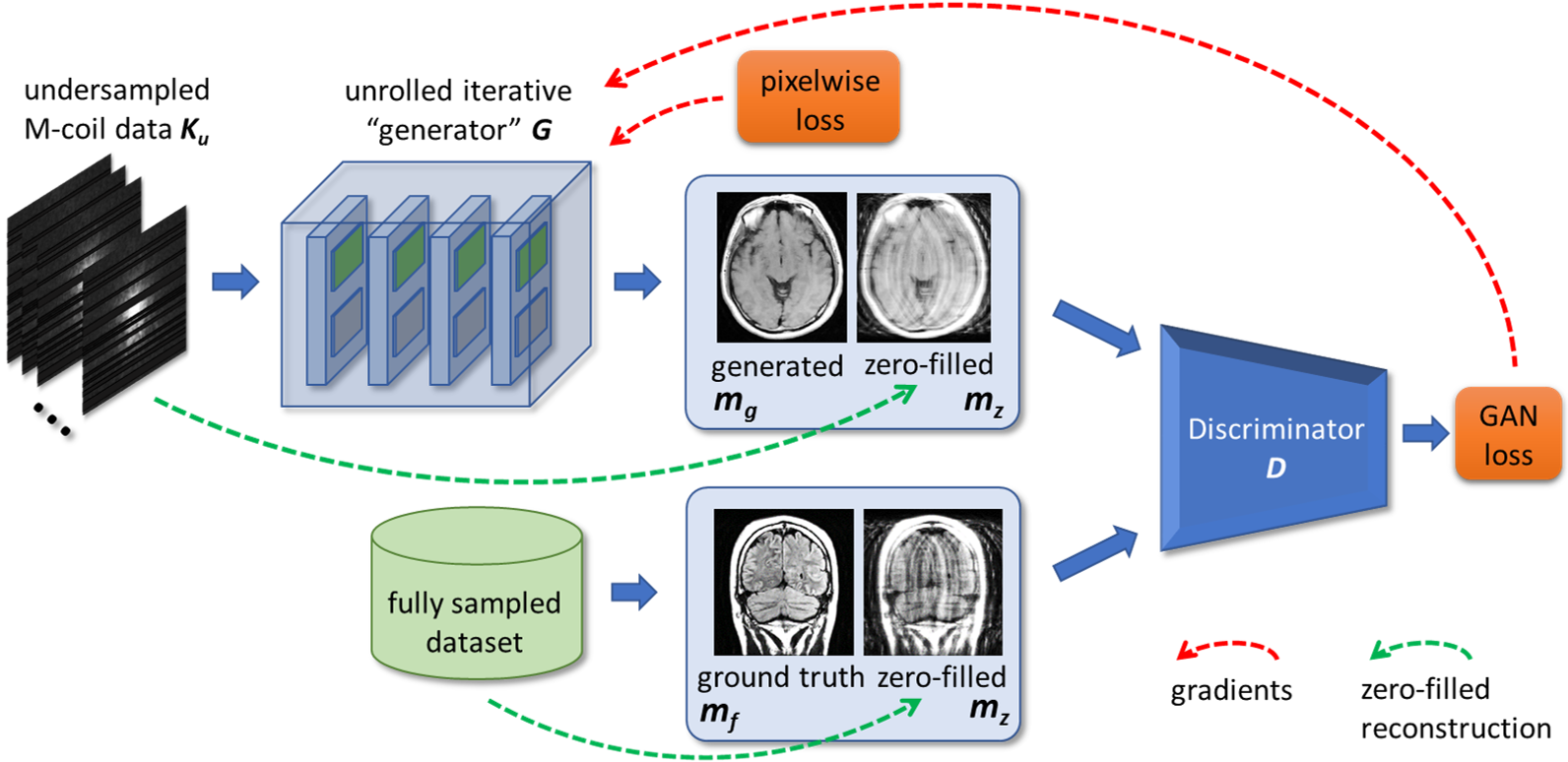}
\centering
\caption{The generator receives undersampled $k$-space data as input and generates a matched estimated fully-sampled image. The discriminator learns to estimate the Wasserstein Distance between “fake” pairs and “real” pairs. 
}
\label{fig:cgan}
\end{figure}

Our method learns a DL-based sparse MRI reconstruction model from training samples, each of which is a pair of a fully sampled and matched undersampled $k$-space data. We propose a conditional GAN architecture, which conditions the reconstruction using the  zero-filled image. Specifically, our model is composed of a generator and a discriminator networks. The generator reconstructs an image from an undersampled $k$-space dataset. The discriminator receives a pair of input images: (i) a ground truth image $m_f$ or a generated (``fake'') reconstructed image from undersampled $k$-space and (ii) a zero-filled image $m_z$ (see Fig.~\ref{fig:cgan}). 

While it is possible to use a non-conditional GAN architecture, in this case the discriminator can only enforce general style properties learned from the distribution of the fully sampled images, and for a given undersampled $k$-space signal, $K_u$, it is not guaranteed that the generator would learn to reconstruct a realistic image $G(K_u)$ that perceptually matches its corresponding $m_z$. 

\noindent{\bf Objective~~}
\label{sec:Objective}
Following the success of the Wasserstein GAN (WGAN)~\cite{29arjovsky2017wasserstein} and the framework proposed by Isola et al.~\cite{20isola2017image}, we adopt a conditional WGAN objective:
\begin{equation}
\pazocal{L}_{cWGAN}(G,D) = \E_{(m_z, m_f)}[D(m_z,m_f)] - \E_{(m_z^k, K_u)}[D(m_z^k,G(K_u))] 
\end{equation}
where $G$ and $D$ are the generator and discriminator networks, repectively. $K_u$ is a random undersampled $k$-space data, $m_f$ is a random fully sampled image, and their corresponding undersampled zero-filled images are  $m_z^k$ and $m_z$ respectively. In addition to the adversarial loss, we also add a pixel-wise Mean Square Error (MSE) loss  $\pazocal{L}_{MSE}(G)= \frac{1}{WH}\sum_{i=1}^{W}\sum_{j=1}^{H}((m_f)_{i,j}-G(K_u)_{i,j})^2$, where W and H are the width and height of the image $m_f$. The final generator loss $\pazocal{L}_{G}$ is: 
\begin{equation}
\pazocal{L}_{G} = \arg\min_G \max_D\pazocal{L}_{cWGAN}(G,D) +\lambda\pazocal{L}_{MSE}(G).
\end{equation}

\noindent{\bf Adaptive Gradient Balancing~~}
In WGAN training, the discriminator network is used as a learned loss function, which dynamically changes during training, and thus may generate gradients with variable norm. To stabilize the WGAN training and to avoid drifting away from the ground-truth spatial information, we introduce the Adaptive Gradient Balancing (AGB) algorithm for continually balancing the gradients of the pixel-wise and the WGAN loss functions. 

In order to keep the gradients of both terms at the same level, and since the WGAN gradients tend to vary, we choose to adaptively upper-bound the WGAN gradients. Specifically, we define $\beta$ to be an adaptive weight that will be used to bound the WGAN loss gradients. We calculate two moving-average variables $g_{ma}$ and $p_{ma}$ corresponding to the WGAN loss and the pixel-wise loss, respectively. These moving averages capture the standard deviation (STD) of the gradients calculated at every backward step on the generated image, with respect to each one of the losses separately. At every training step, if
$g_{ma} > p_{ma} \cdot ratio$
for a predefined $ratio$ value, we update $g_{ma}$ and $\beta$ as follows:
$\beta \gets \beta \cdot (1 + rate)$,
$g_{ma} \gets g_{ma} \cdot (1 - rate)$,
where $rate$ is a predefined decay rate. During training, we divide the WGAN loss by $\beta$ to carefully decay the WGAN loss gradients to roughly the same order of magnitude as those of the pixel-wise loss. Moreover, in order to keep a reasonable ratio between the generator's WGAN loss gradients and the discriminator loss gradients, we also decay the discriminator loss by the same $\beta$ factor (see Alg.~1).  

\begin{algorithm}[t] 
  \caption{WGAN-AGB training of WGANs. Parameters: $\alpha = 5 \cdot 10^{-5}$, $\beta_{init} = 10$, $c = 0.01$, $\lambda = 0.99$, $ratio = 10$, $rate = 0.1$, $n_{discriminator} = 1$ }
  \begin{multicols}{2}
    \begin{algorithmic}
      \scriptsize
      \STATE  $p_{ma} \gets 0$; $g_{ma} \gets 0$; $\beta \gets \beta_{init}$;
      \FOR{number of training iterations} {
        \FOR{$t$ = 0, ..., $n_{discriminator}$} {
            \STATE Sample a minibatch \{($x^{i}, z^{i}$)\}$_{i=1}^{m}$ 
            \STATE $g_w$ $\gets$ $\nabla _w$ $[\frac{1}{m}\frac{1}{\beta}$ $(\sum_{i=1}^{m} D_w(x^{i})$ - \par \hskip\algorithmicindent $\sum_{i=1}^{m} D_w(G_\theta(z^{i})))]$  \;
            
            \STATE $w$ $\gets$ $w$ + $\alpha$ $\cdot$ Adam($w$, $g_w$) \;
            \STATE $w$ $\gets$ clip($w$, -$c$, $c$)
            
            }\ENDFOR
            
        \STATE Sample a minibatch \{($x^{i}, z^{i}$)\}$_{i=1}^{m}$ 
        
        \STATE $g_\theta$ $\gets$  $  \nabla _\theta$ $ [$ $\frac{1}{m}(\frac{1}{\beta}\sum_{i=1}^{m} D_w(G_\theta(z^{i}))$ + 
        \par \hskip\algorithmicindent$ \sum_{i=1}^{m} MSE(x^{i}, G_\theta(z^{i})))]$ \;

    \columnbreak
      \scriptsize
        \STATE $\theta$ $\gets$ $\theta + \alpha$ $\cdot$ Adam($\theta$, $g_\theta$) \;
       \STATE $g_{gan}$ $\gets$ $\frac{1}{m}\frac{1}{\beta}\sum_{i=1}^{m}  \nabla _{G_\theta(z^{i})} [D_w(G_\theta(z^{i})]$\;
       \STATE $g_{MSE}$ $\gets$ $ \frac{1}{m}\sum_{i=1}^{m}  \nabla _{G_\theta(z^{i})} [MSE(x^{i}, G_\theta(z^{i}))]$\;
       \STATE $g_{ma} \gets g_{ma} \cdot \lambda + (1-\lambda) \cdot STD(g_{gan})$\;
       \STATE $p_{ma} \gets p_{ma} \cdot \lambda + (1-\lambda) \cdot STD(g_{MSE})$\;
       \IF{ $g_{ma} > p_{ma} \cdot ratio $}{
       \STATE $\beta \gets \beta \cdot (1 + rate)$ \; 
       \STATE $g_{ma} \gets g_{ma} \cdot (1 - rate)$
       }\ENDIF
    } \ENDFOR
    \end{algorithmic}
  \end{multicols}
\end{algorithm}

Our AGB algorithm extends WGAN training and ensures one invariant during the entire training - the STD of the WGAN loss gradients is upper-bounded by a factor of the STD of the pixel-wise loss gradients. This invariant maintains the effectiveness of both loss terms, over the entire course of training.

\begin{figure*}[t]
  \includegraphics[width=0.8\linewidth]{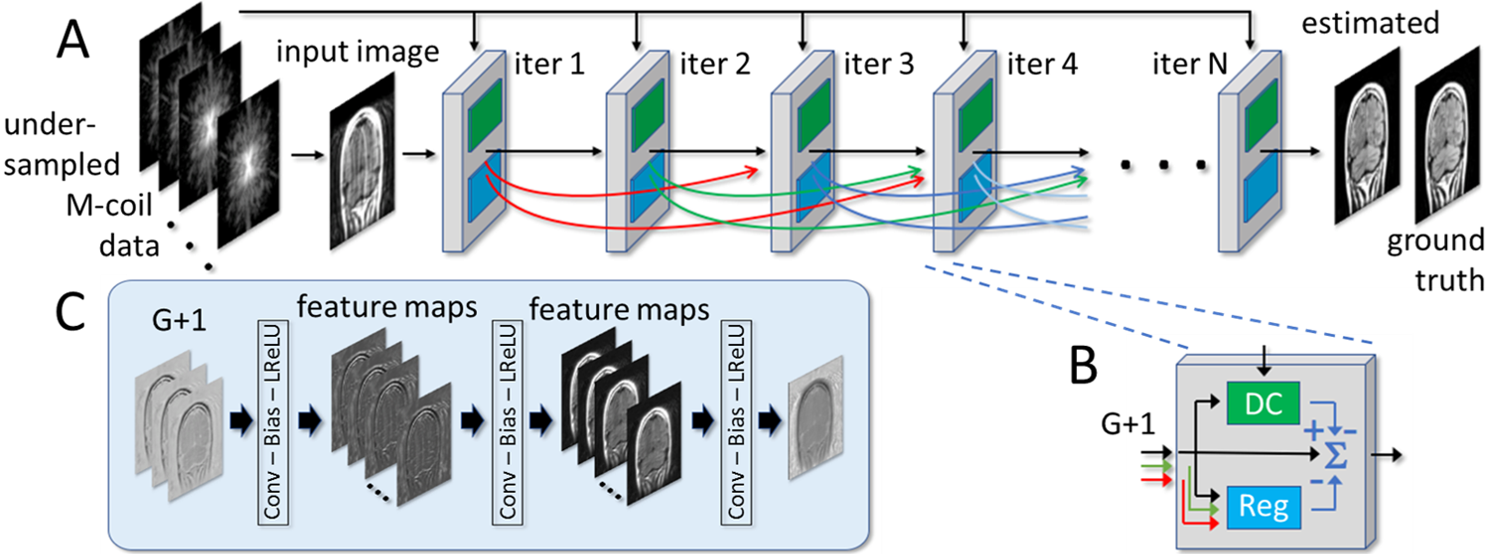}
  \centering
  \caption{DCI-Net (A) consists of N unrolled iterative blocks, each with dense skip-layer connections (curved arrows) to subsequent blocks. Each iterative block (B) consists of data-consistency (DC) and convolutional (C) units. The convolutional unit operates on all G+1 connections, while DC units operates only on direct connection. 
  }
    \label{fig:dci}
\end{figure*}

\noindent{\bf Network architectures~~}
We propose a new generator architecture (Fig.~\ref{fig:dci}), called Densely Connected Iterative Network (DCI-Net), which is based on the iterative convolutional network \cite{8hammernik2018learning,9schlemper2018deep}. 
The key new developments are the use of (1) dense connections~\cite{10huang2017densely} across all iterations, which strengthens feature propagation, making the network more robust, and (2) a relatively deep architecture of over 60 convolutional layers, bringing increased capacity.  Our generator receives M coils of undersampled $k$-space data, and uses N = 20 iterations, each of which includes a data-consistency unit and a convolutional unit for regularization (Fig.~\ref{fig:dci}B). Dense skip-layer connections between the output of each iteration and the following G iterations -- where typically G = 5 -- are represented as curved lines in Fig.~\ref{fig:dci}A. This results in an input to each block composed of skip and direct connections concatenated to form a G+1 channel complex image. For our discriminator architecture we use a convolutional ``PatchGAN''~\cite{18li2016precomputed}. More information 
can be found in Appendix~\ref{sup:architecture}.

\section{Results}
\noindent{\bf Dataset~~}
Fully sampled brain MRI datasets (T1, T2, T1-FLAIR and T2-FLAIR in axial, coronal and sagittal orientations) were acquired with various k-space data sizes and various numbers of coils along with sensitivity maps estimated from separate calibration scans. In total, 2267 slices were acquired, of which 1901 were used to train the networks, 151 for validation and 215 for testing. In addition, during training, we also applied random horizontal flips and rotations (bounded to 20 degrees) to augment the training set. The data were retrospectively down-sampled using 12 central lines of k-space and a 1D variable-density sampling pattern outside the central region, resulting in a net under-sampling factor $R = 4$. As evaluation metrics, we compute both normalized mean square error (NMSE), and the Fr\'echet Inception Distance (FID)~\cite{30heusel2017gans}, which is a similarity measure between two datasets that correlates well with human judgment of visual quality and is most often used to evaluate the quality of images generated by GANs. The Adam optimizer is used with a learning rate of 5x$10^{-4}$ for both generator and discriminator networks. For the traditional GAN training, $\lambda$ is initialized to 100, after a hyper parameter search conducted on the values 10, 100, 1000. All models performed 600 epochs in $\sim$2 weeks of training, and the inference run time is 100ms per slice on a single GPU.

\begin{table}[t]
\begin{minipage}[c]{0.42\linewidth}
\caption{Comparison of our method with zero-filled images (ZF), and reconstruction using wavelets or total variation (TV)~\cite{32TV_WAVELETS_lustig2007sparse} and ARC~\cite{31ARC_beatty2007method}. NMSE is w.r.t fully sampled image.} 
\label{table:comprison-to-classic-methods}
\centering
\begin{tabular}{|l|c|c|}
\hline
Images & NMSE & FID \\
\hline\hline
ZF & 115 &  173.0\\
Wavelets & 18.7 &  138.4\\
TV & 14.1 &  117.0\\
ARC & 18.9 &  109.0\\
cWGAN-AGB   & \textbf{3.39}  &  \textbf{18.7}\\ 
\hline
\end{tabular}
\end{minipage}%
\hfill
\begin{minipage}[c]{0.56\linewidth}
\caption{Evaluation on a holdout test set. The WGAN variants all employ a generator with 20 iterations (20I), a growth rate of 5 (5G) and 40 kernels for each convolution (40K). 
}  
\label{table:quanitative-results}
\centering
\begin{tabular}{|l|c|c|}
\hline
Experiment & NMSE  & FID\\
\hline\hline
DCI-Net (5I-5G-160K) & 3.67 & 20.2 \\ 
DCI-Net (20I-1G-40K, no dense) & 3.46 & 19.3\\
DCI-Net (20I-5G-40K) & \underline{3.24} & 19.4\\
\hline
WGAN & 3.71  & 19.7 \\
cWGAN & 3.61 &  19.9\\
cWGAN-AGB (proposed) & \textbf{3.39} &  \textbf{18.7}\\
\hline
\end{tabular}
\end{minipage}
\smallskip
\smallskip
\caption{Mean of sharpness, SNR, contrast, artifacts and overall IQ scored for our proposed cWGAN-AGB, a baseline DCI-Net and the fully-sampled images. Scores 1 to 5 indicate poor to excellent.
}
\label{table:blind-test}
\centering
\begin{tabular}{|l|c|c|c|c|c|c|}
\hline
Images & Sharpness & SNR & Contrast & Artifacts & Overall IQ\\
\hline\hline
Fully sampled & 5.0 & 3.3 & 4.0 & 4.0 & 4.5 \\
Baseline (DCI-Net) & 2.3 & 4.5 & 4.0 & 3.8 & 2.3 \\
cWGAN-AGB (proposed) & 3.8 & 3.8 & 4.0 & 3.8 & 3.5 \\
\hline
\end{tabular}
\end{table}

\noindent{\bf Comparison with baseline methods~~}
We compare on the test set our cWGAN-AGB to compressed sensing methods using wavelets or Total Variation (TV) ~\cite{32TV_WAVELETS_lustig2007sparse} and to Autocalibrating Reconstruction for Cartesian imaging (ARC) ~\cite{31ARC_beatty2007method}. As can be seen in Tab.~\ref{table:comprison-to-classic-methods}, our proposed model produces significantly more accurate reconstructions than the other methods. 

\begin{figure*}[t]
\includegraphics[width=0.9\linewidth]{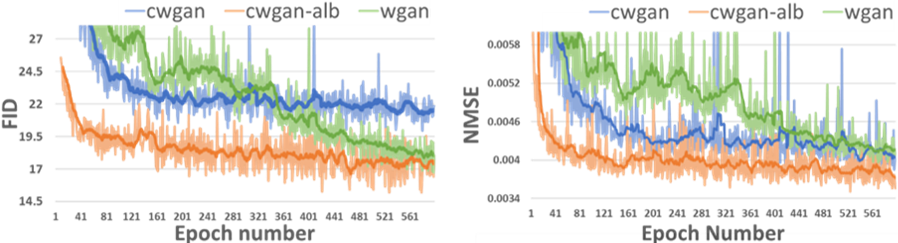}
\centering
\caption{FID and NMSE (lower is better) during training, as evaluated on the validation set. Results are shown for WGAN, a vanilla cWGAN, and our adative cWGAN-AGB.}
\label{fig:val-curve}
\end{figure*}

\noindent{\bf Comparing GANs convergance~~}
To show the effectiveness of our method, we compared the convergence of our cWGAN-AGB model to those of cWGAN and WGAN, trained without AGB. During the training phase, FID and NMSE were evaluated on a hold-out validation set, for each epoch. As can be seen in Fig.~\ref{fig:val-curve}, our proposed model converges better, with both scores decreasing significantly faster compared to the other techniques. 

\begin{figure*}[t]
\includegraphics[width=0.7\linewidth]{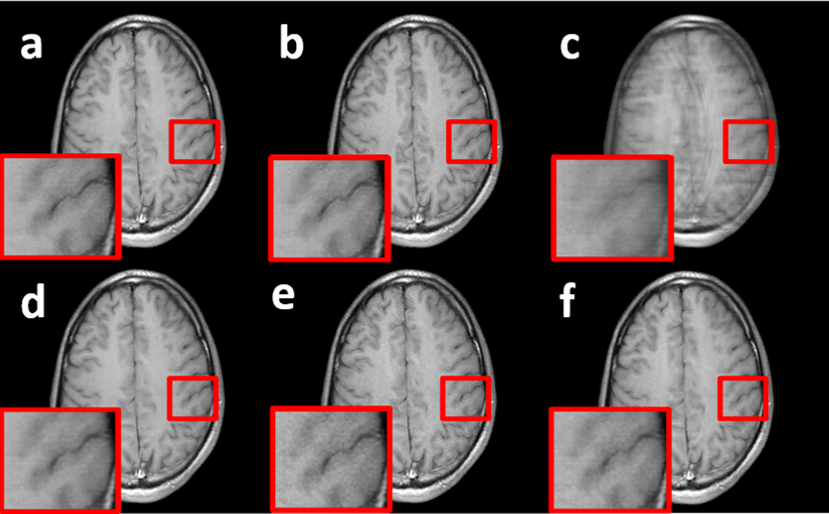}
\centering
\caption{A representative example with regions of interest showing the reconstruction of all models side-by-side: a) cWGAN-AGB; b) ground-truth (fully-sampled) image; c) zero-filled image; d) baseline generator network; e) WGAN; f) cWGAN. cWGAN and cWGAN-AGB have better SNR and fewer artifacts than WGAN. cWGAN-AGB yields sharper images with more fine details while maintaining a more natural appearance. The baseline model sometimes exhibits some blurring.}
\label{fig:imageCompare}
\end{figure*}

\noindent{\bf Ablation analysis~~}
We compare, in Tab.~\ref{table:quanitative-results}, our cWGAN-AGB with 3 other models: 1) cWGAN, 2) WGAN, and 3) a baseline DCI-Net for sparse MRI reconstruction without any GAN technique. All models were evaluated with NMSE and FID on the test set. We found that (a) cWGAN and cWGAN-AGB have better SNR and fewer artifacts than WGAN, (b) cWGAN-AGB converges much faster than cWGAN and WGAN (see Fig.~\ref{fig:val-curve}) and performs better in both FID and NMSE measures (Tab.~\ref{table:quanitative-results}) and (c) although cWGAN-AGB has higher NMSE than the baseline model, it performs better in FID and yields sharper images with more fine details while maintaining a natural image texture (see Fig.~\ref{fig:imageCompare}).

In Tab.~\ref{table:quanitative-results}, we also compare to baseline architectures, 
demonstrating the effectiveness of our key new  architecture developments: (1) dense connections across all iterations, which strengthens feature propagation, making the network more robust, and (2) a relatively deep architecture of 20 iterations, composing more than 60 convolutional layers, which brings an increased capacity. We compared our generator to (1) a similar network without dense connections and (2) a 5-iteration based network with a similar number of learned parameters. Employing dense connections significantly improved accuracy, and the use of the deeper network produced 12\% lower mean NMSE than a shallower network that had a similar number of learned parameters.

\noindent{\bf Visual Scoring~~}
To assess the perceptual quality of the resulting images we report a visual scoring conducted by four experienced MRI scientists. The same test set was ranked for cWGAN-AGB, the baseline method and for the fully sampled images. The scoring was performed blindly and the images were randomly shuffled. The studies were taken from a cohort of seven healthy volunteers. Each study contained a full brain scan comprising 25-43 slices. For each study, image sharpness, signal-to-noise ratio (SNR), contrast, artifacts and overall image quality (IQ) were reported. Tab.~\ref{table:blind-test} shows that cWGAN-AGB produced significantly sharper images than the baseline network, at the cost of somewhat weaker denoising of the images.



\section{Conclusions}
We present a novel sparse MRI reconstruction model that employs a cWGAN loss term, and a novel GAN training procedure. By leveraging GANs to their fullest, the method generates sharper images with more fine detail and natural appearance than would otherwise be possible. In addition, dense connections are used to improve the performance of our unrolled iterative generator network. 
In the context of MRI reconstruction, a GAN based model can raise concerns about hallucination, where image details that do not appear in the ground truth are generated. We found that our method produces significantly less hallucination than other GANs. This may be due to the usage of (1) a pixel-wise loss term, that prioritizes reconstruction accuracy, (2) data consistency layers embedded inside the network, (3) a conditional GAN architecture that allows the discriminator to penalize low-fidelity reconstruction and (4) our AGB training, that continuously upper bounds the gradients of the GAN loss. Moreover, we believe our AGB training can be beneficial for any GAN-based model employing a multi-term loss objective, especially in the medical domain where there is more variability in the input and less experience in balancing GAN loss terms.

%
%
\bibliographystyle{splncs04}
\bibliography{bibshort}


\appendix

\section{Network architectures} \label{sup:architecture}

\subsection{Data-consistency unit}
\begin{figure*}[h]
  \includegraphics[width=\linewidth]{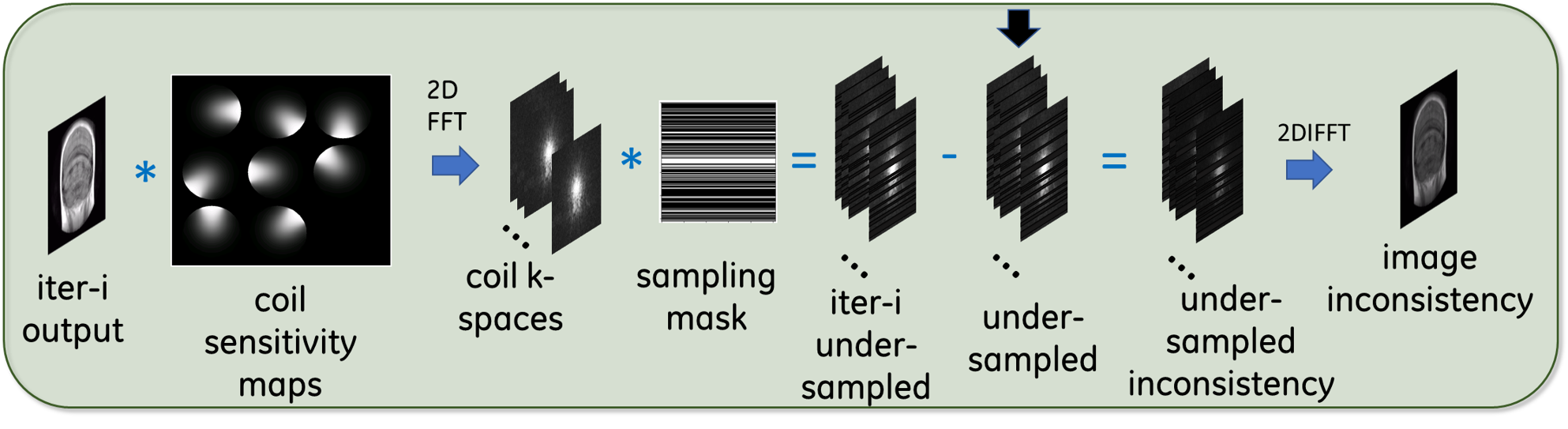}
  \centering
  \caption{Data Consistency (DC) unit. Each iteration contains a DC unit that operates only on the iteration's direct input image. The DC calculates the inconsistencies between (i) the undersampled $k$-space of the iteration's input image and (ii) the acquired $k$-space. By using Inverse Fourier Transform, the calculated inconsistency is transformed to image space, then multiplied by a learned weight and subtracted from the iteration's input image (not shown in the figure).}
    \label{fig:dataConsistency}
\end{figure*}

Each data-consistency unit shades the input image with each coil sensitivity map, transforms the resulting images to $k$-space, imposes the sampling mask, calculates the difference relative to acquired $k$-space and returns them to the image domain, multiplied by a learned weight (Fig.~\ref{fig:dataConsistency}). By utilizing the acquired $k$-space data as a prior, the data-consistency units, embedded as operations inside the network, keep the network from drifting away from the acquired data. For this use, the undersampled $k$-space data were also input directly into each iterative block of the network (Fig. 2A,B of the main text).

\subsection{Convolutional unit}
Each convolutional unit (Fig. 2C of the main text) has three sequences consisting of 5x5 convolution, bias, and leakyReLU \cite{36_lrelu_xu2015empirical} layers. The output of the final iteration (Fig. 2A of the main text) is (1) compared to the fully sampled reference image to generate a pixel-wise loss function, using MSE, and (2) paired with its corresponding zero-filled image and fed into a discriminator network to evaluate WGAN ~\cite{29arjovsky2017wasserstein} loss. 

\subsection{discriminator architecture}

For our discriminator architecture we use a convolutional ``PatchGAN'' \cite{18li2016precomputed}. The discriminator receives a pair of (1) $m_z$ and (2) $m_f$ or $G(K_u)$,  concatenated as two channels and is able to penalize structure at the scale of image patches, from both channels. The architecture incorporates four convolutional layers with a stride of 2, each followed by batch normalization \cite{35_batchNorm_ioffe2015batch} and LeakyReLU \cite{36_lrelu_xu2015empirical}. The last convolutional layer is flattened and then fed into a linear layer, for which each input value corresponds to a different patch in the input channels. The linear layer outputs a single value, which is used to calculate the discriminator's WGAN loss. 

\section{Results} \label{sup:results}

\subsection{Qualitative results}
\begin{figure*}[t]
  \includegraphics[width=\linewidth]{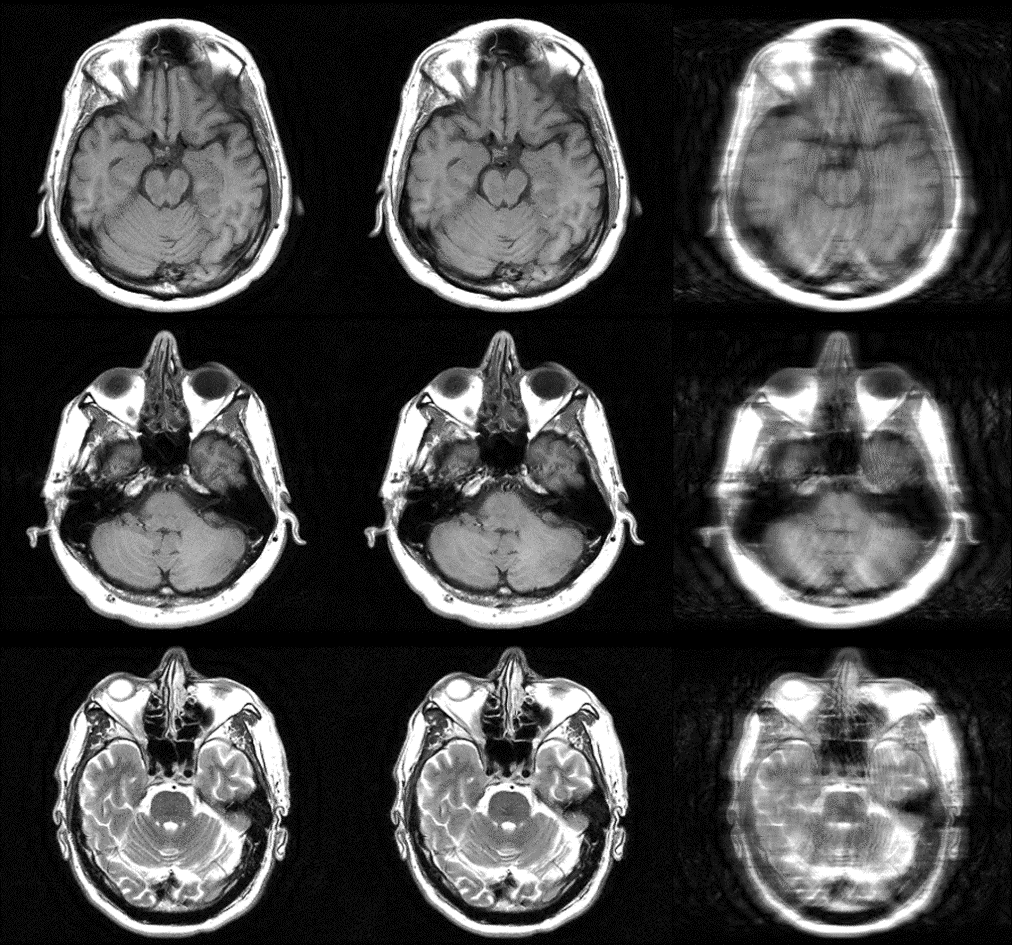}
  \centering
  \caption{ More results of our approach. Left to right: fully sampled, proposed method, zero filled images (R=4).
  }
    \label{fig:qualitative-more}
\end{figure*}

\begin{figure*}[t]
  \includegraphics[width=\linewidth]{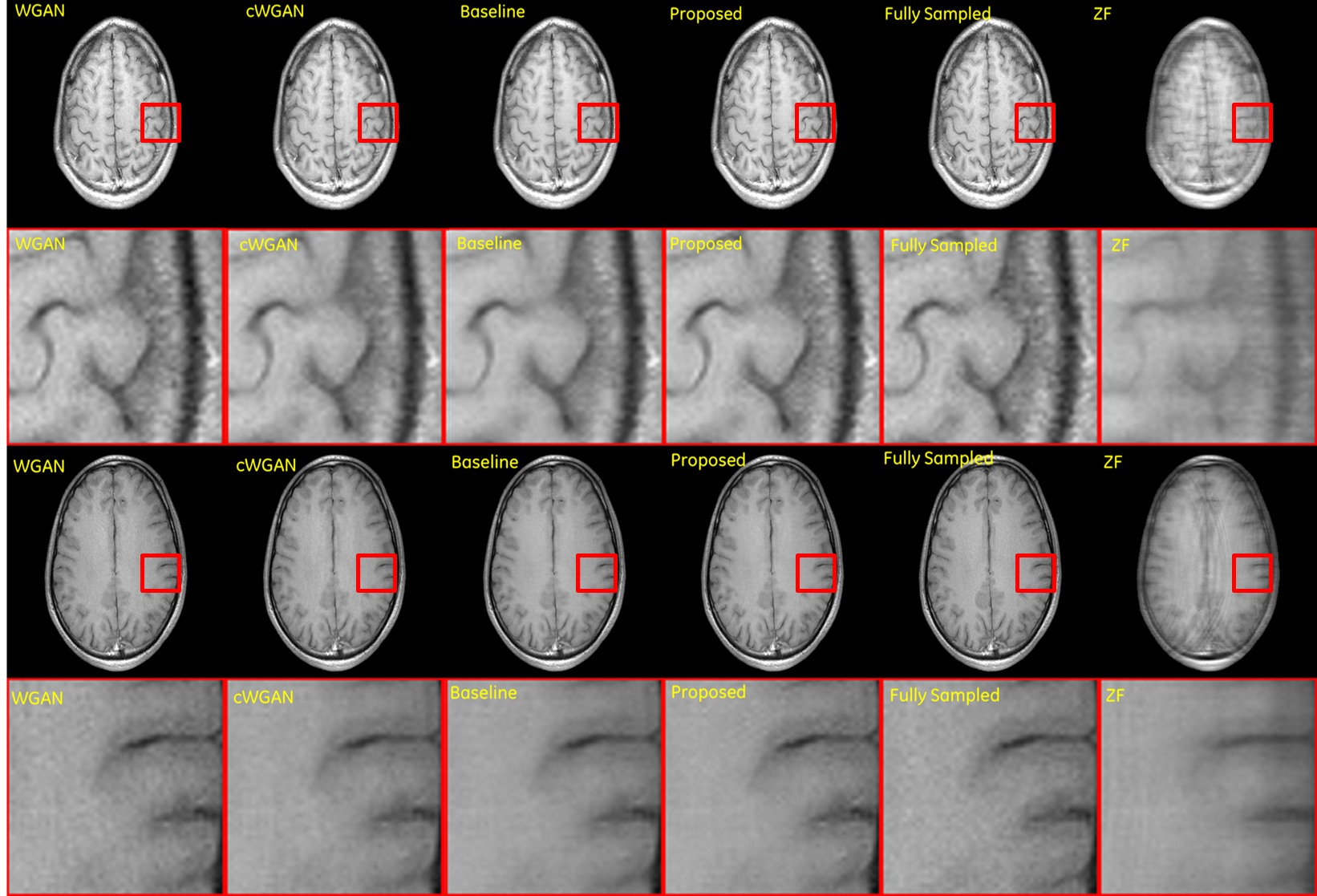}
  \centering
  \caption{More qualitative results from our ablation analysis.
  }
    \label{fig:more-ablation}
\end{figure*}

Fig.~\ref{fig:qualitative-more} exhibits more qualitative results of our proposed model, along with the zero filled (ZF) and the fully sampled images. Fig.~\ref{fig:more-ablation} provides more qualitative results from our ablation study, comparing three different GAN models and a baseline model, where the baseline is our proposed DCI-Net trained solely to optimize MSE loss.

\begin{figure*}[t]
  \includegraphics[width=\linewidth]{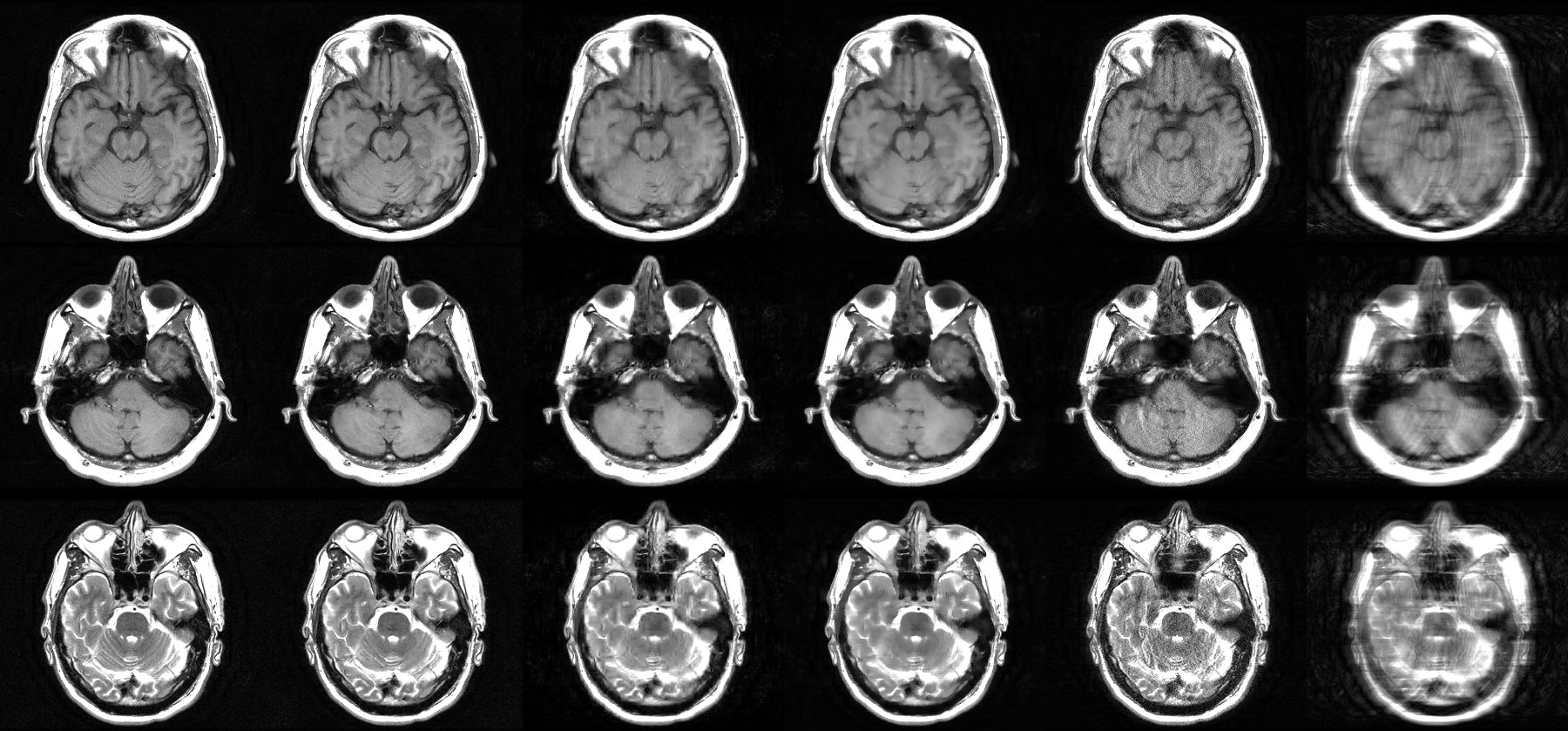}
  \centering
  \caption{Comparison with baseline methods. Left to right: fully sampled, proposed method, wavelets, Total Variation, ARC, zero filled.
  }
    \label{fig:qualitative-classic}
\end{figure*}

For the sake of completeness, we provide a qualitative comparison of our proposed model to compressed sensing methods using wavelets or Total Variation (TV) ~\cite{32TV_WAVELETS_lustig2007sparse} and to Autocalibrating Reconstruction for Cartesian imaging (ARC) ~\cite{31ARC_beatty2007method}, as shown in Fig.~\ref{fig:qualitative-classic}. It can be seen that our proposed method produces higher-quality images than baseline methods, both in terms of perceptual quality and reconstruction error.

\subsection{Implementation Details}
\begin{figure*}[t]
  \includegraphics[width=0.65\linewidth]{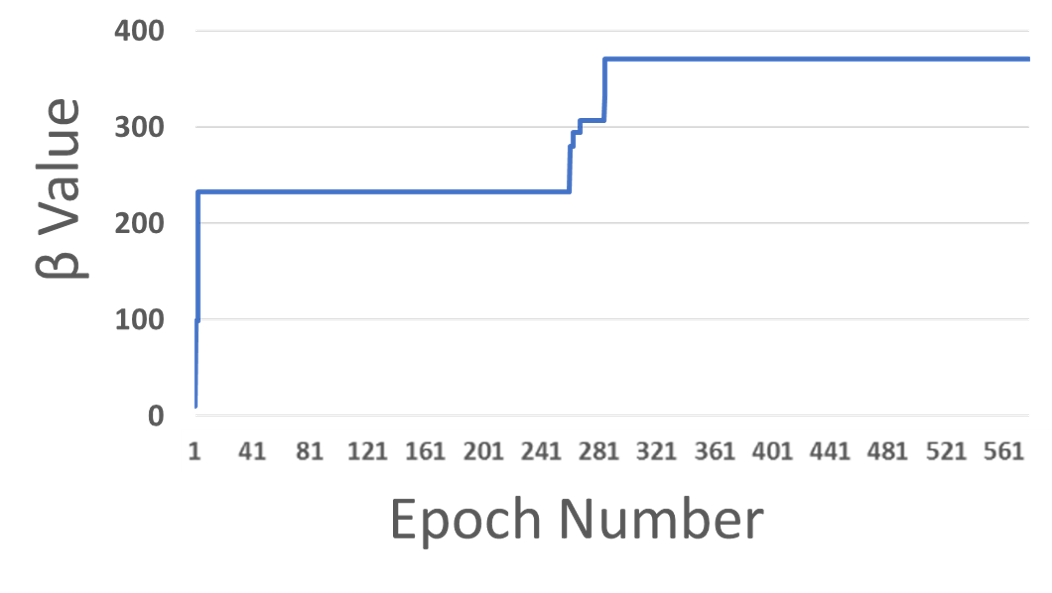}
  \centering
  \caption{Beta value calculated per epoch, for our proposed cWGAN-AGB model. 
  }
    \label{fig:beta}
\end{figure*}

Adam optimizer \cite{34_adam_kingma2014adam} is used with a learning rate of $5\times10^{-4}$ for both generator and discriminator networks, with the momentum parameter $\beta_{m}$ = 0.9. Training is performed with TensorFlow interface on a GeForce GTX TITAN X GPU, 12GB RAM. For the proposed model with AGB training, $\beta$ is initialized to 10 and increased in multiple steps during training to a value of 370 (see Fig.~\ref{fig:beta}).

\subsection{Model Selection} 
In this study, we use both NMSE and FID \cite{30heusel2017gans} for model selection. Specifically, to select the best model for each experiment, we evaluate FID and NMSE on the validation set for each epoch (see Fig. 3 of the main text). Then, we calculate the mean of both scores per experiment (starting from epoch 200), and normalize each series separately, by subtracting and dividing by their corresponding mean and STD, respectively. The epoch for which the model minimizes the sum of normalized FID and normalized NMSE has been selected for evaluation on the test set. 

\subsection{Sampling Pattern} 
\begin{figure}[t]
\includegraphics[width=\linewidth]{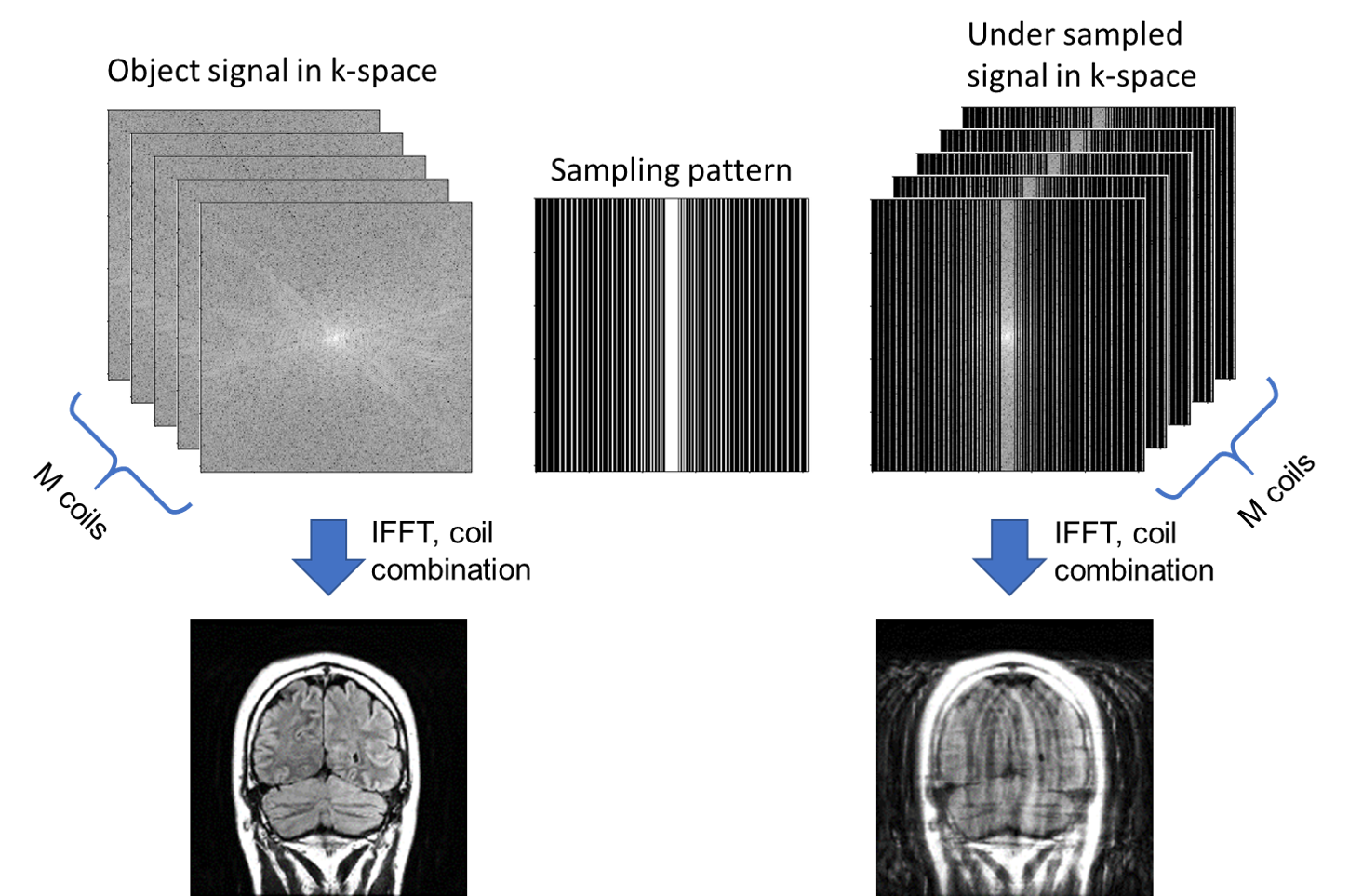}
  \centering

  \caption{Fully-sampled $k$-space multiplied by an acquisition sampling pattern, with acceleration factor of 4, results in highly undersampled $k$-space (top right). Conventional reconstruction of the undersampled $k$-space using zero-filling generates a low-quality image with heavy artifacts that is completely non-diagnostic (bottom right). A non-accelerated acquisition that uses fully-sampled $k$-space results in high-quality image (bottom left). In this study, we focus on 2D data acquisition, which utilizes a 1D sampling pattern in the phase-encoding direction.
  }
  \label{fig:introFig}
    
\end{figure}

Our method is applied to accelerated multi-slice 2D scanning, where $k$-space is undersampled in the phase-encode direction using a 1D variable-density sampling (VDS) pattern ~\cite{33_VDS_tsai2000reduced} whose density decreases linearly between the central and outer regions of $k$-space (with a net undersampling factor of four), but with the central 12 lines of $k$-space fully sampled (see Fig.~\ref{fig:introFig}). 

\end{document}